\documentclass[letterpaper, 10 pt, conference]{ieeeconf}  
\IEEEoverridecommandlockouts                          
\usepackage{cite}
\usepackage{amsmath}
\usepackage{graphicx}
\usepackage{color}
\usepackage{multirow}
\usepackage{gensymb}
\usepackage{balance}
\usepackage[caption=false,font=footnotesize]{subfig}
\usepackage{algorithm}
\usepackage{algorithmic}

\title{\LARGE \bf
Motion Planning for a UAV with a Straight or Kinked Tether}

\author{Xuesu Xiao$^{1}$, Jan Dufek$^{1}$, Mohamed Suhail$^{2}$ and Robin Murphy$^{1}$
\thanks{$^{1}$Xuesu Xiao, Jan Dufek, and Robin Murphy are with the Department of Computer Science and Engineering,
        Texas A\&M University, College Station, TX 77843
        {\tt\small \{xiaoxuesu, dufek, robin.r.murphy\}@tamu.edu}}%
\thanks{$^{2}$Mohamed Suhail is with the Department of Visualization, Texas A\&M University, College Station, Texas 77843
        {\tt\small mohamedsuhail@tamu.edu}}%
}

\begin{document}

\maketitle
\thispagestyle{empty}
\pagestyle{empty}

\begin{abstract}

This paper develops and compares two motion planning algorithms for a tethered UAV with and without the possibility of the tether contacting the confined and cluttered environment. 
Tethered aerial vehicles have been studied due to their advantages such as power duration, stability, and safety. However, the disadvantages brought in by the extra tether have not been well investigated by the robotic locomotion community, especially when the tethered agent is  locomoting in a non-free space occupied with obstacles. 
In this work, we propose two motion planning frameworks that (1) reduce the reachable configuration space by taking into account the tether and (2) deliberately plan (and relax) the contact point(s) of the tether with the environment and enable an equivalent reachable configuration space as the non-tethered counterpart would have. Both methods are tested on a physical robot, Fotokite Pro. 
With our approaches, tethered aerial vehicles could find their applications in confined and cluttered environments with obstacles as opposed to ideal free space, while still maintaining the advantages from the usage of a tether. The motion planning strategies are particularly suitable for marsupial heterogeneous robotic teams, such as visual servoing/assisting for another mobile, tele-operated primary robot. 

\end{abstract}

\section{INTRODUCTION}
\begin{figure}[!t]
\centering
\subfloat[First Person View of \newline Robot's Onboard Camera]{\includegraphics[width=0.44\columnwidth]{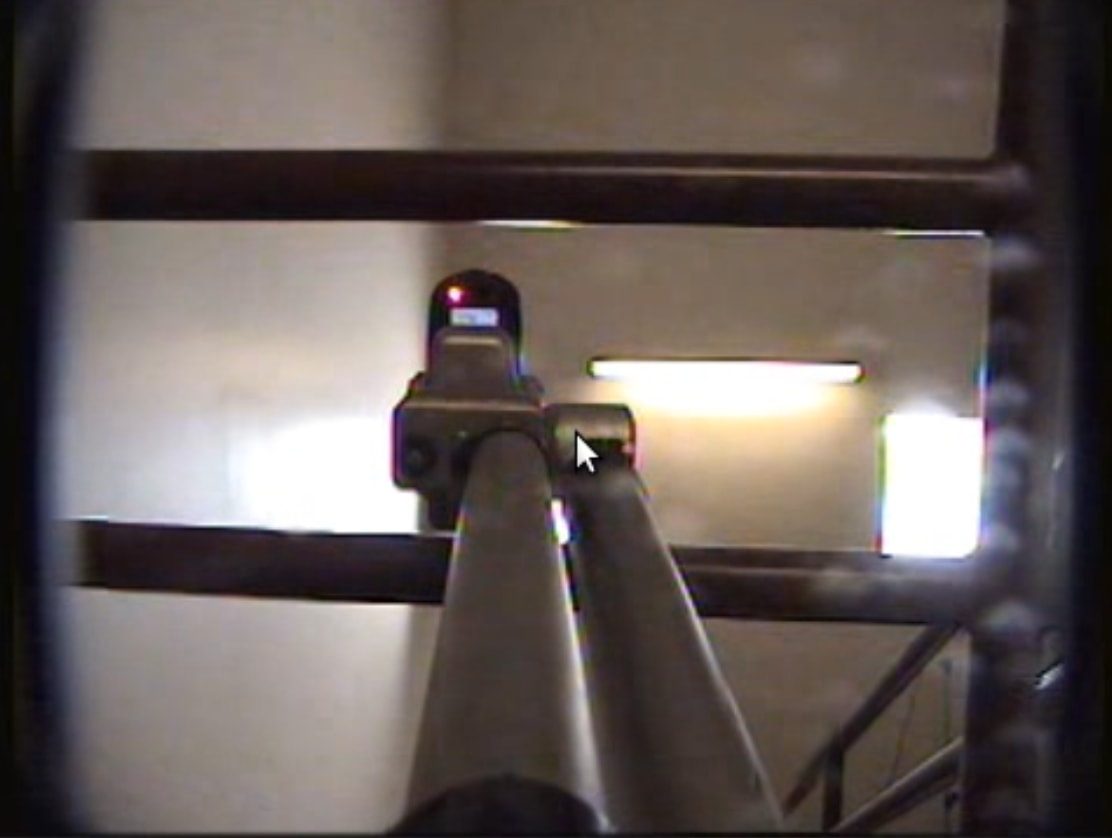}%
\label{fig::fpv}}
\subfloat[Third Person View from Assistive Robot Helping Primary Robot Open a Door]{\includegraphics[width=0.55\columnwidth]{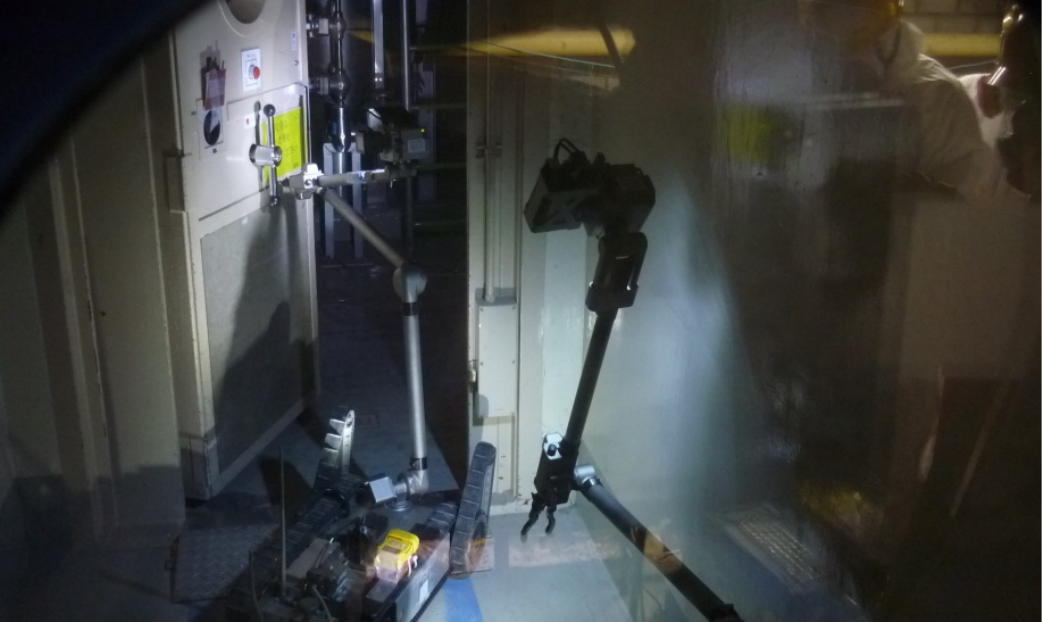}%
\label{fig::2packbots}}
\hfil
\subfloat[Heterogeneous Marsupial Robot Team with Autonomous Visual Assistant Using a Tethered UAV]{\includegraphics[width=0.8\columnwidth]{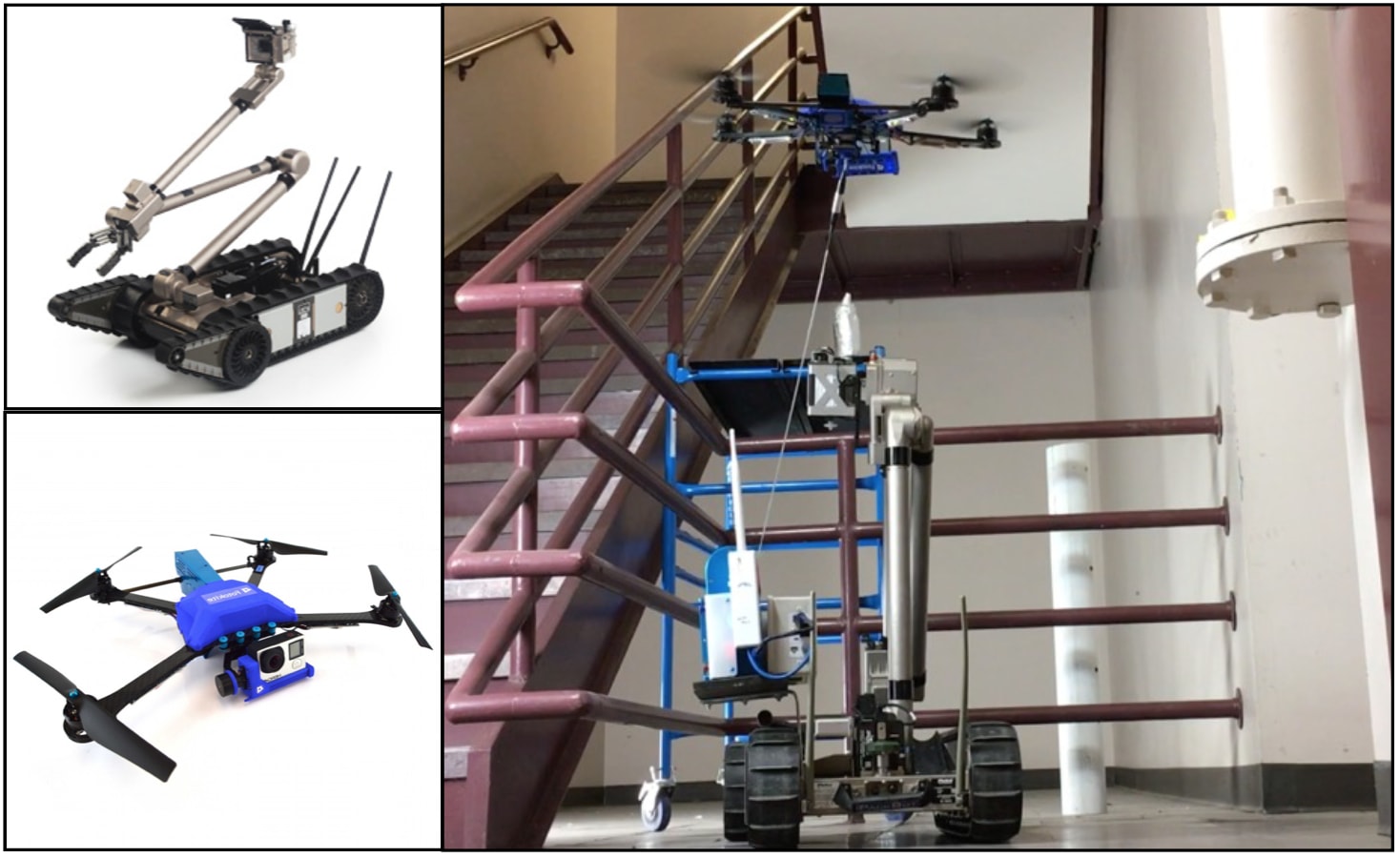}%
\label{fig::team}}
\caption{Teleoperation solution in a remote environment is improved from first person view by onboard camera (a), through a separate manually controlled ground robot for third person view (b), to an autonomous flying aerial visual assistant operating with a tether (c) \cite{xiao2017visual}. }
\label{fig::solutions}
\end{figure}

Tethered Unmanned Aerial Vehicles (UAVs) have been extensively investigated in the literature due to their alternative advantages when being compared to their non-tethered counterparts, such as stability \cite{lupashin2013stabilization}, agility \cite{schulz2015high}, and extended flight duration \cite{zikou2015power}. They also provide extra safety margins and are in accordance with certain authority requirements \cite{pratt2008use, murphy2016two}. Due to such advantages, tethered UAVs have also been paired with Unmanned Ground Vehicles (UGVs) \cite{xiao2017visual} or Unmanned Surface Vehicles (USVs) \cite{dufek2017visual, xiao2017uav} to increase the primary robot's or the human operator's situational awareness. In particular, a tethered UAV was used to replace tele-operated visual assistant to provide primary robot's operator with better third-person visual feedback and thus improved situational awareness (Fig. \ref{fig::solutions}).

In remote confined and cluttered environments, teleoperation of a robot with first person view from robot's onboard camera is difficult due to the perception limitations, such as the lack of depth perception (Fig. \ref{fig::fpv}). Using a separate tele-operated robot can partially solve this problem by providing more informative additional view points. For example, Fig. \ref{fig::2packbots} shows two iRobot PackBots being used to conduct radiation surveys and read dials inside the plant facility, where the second PackBot provides camera views of the first robot in order to manipulate door handles, valves, and sensors faster. However, this approach introduces other problems including extra 1-2 human operators for the second robot and difficulty in communication between the two operator crews. Therefore, researchers started looking into replacing the second robot and its operating crew with an autonomous agent, especially UAVs due to their superior mobility to cover larger spaces (Fig. \ref{fig::team}) \cite{xiao2017visual}. Moreover, adding a tether to the UAV can provide extra benefits to the visual assisting approach, including matching battery duration with the primary ground robot, reliable localization with light computational overhead, and easy retrieval in the case of malfunction or accident. This requires the autonomous tethered UAV to be able to plan an executable path and navigate through the cluttered environments, while being connected by a tether.

However, planning the motion of a tethered UAV has not been deeply researched, especially in a non-free environment with obstacles. This paper investigates how to generate executable motion plans so that the tethered UAV can navigate through cluttered environments with obstacles. Two motion planners are proposed to allow collision-free motion of the visual assistant as if it were tetherless, while still utilizing the advantages of being tethered to its primary robot. 

This paper is organized as follows: Sec. \ref{sec::related_work} reviews related work regarding motion planning for tethered robots. Sec. \ref{sec::approach} presents two motion planning algorithms: (1) reachable space reduction via ray casting and (2) contact point(s) planning and relaxation, along with their motion executor. Sec. \ref{sec::experiments} gives results from exploratory trials using a physical robot. Discussion on results is presented in Sec. \ref{sec::discussions}. Sec. \ref{sec::conclusions} concludes the paper.

\section{RELATED WORK}
\label{sec::related_work}
Tether planning is not deeply investigated by the Unmanned Aerial Vehicle (UAV) community. \cite{zikou2015power} presented a tethered UAV platform with the purpose of power-over-tether considerations with a specific tether lengthening and retraction method. However, other than power considerations the tethered UAV was not treated differently as a tetherless agent. \cite{schulz2015high} used a tether to achieve high-speed steady flight on a UAV. Novel tether management system and dynamics have been propsed in \cite{minor2002automated, prabhakar2005dynamics}. All these researches focused on hardwares and controls, not tether planning. 

For Unmanned Ground Vehicles (UGVs) and Unmanned Marine Vehicles (UMVs), tether has been widely used, but there is also very limited reported literature on planning with tethers. Tethers were found in robotic missions since tethers can mitigate the constraints imposed by onboard batteries, provide more reliable communication, enable transportation of gases, fluids, and other materials,  and serve as a failsafe to retrieve the robot in case of malfunction \cite{krishna1997tethering, marques2007search}. Tether systems were also designed to provide extra support or to rappel Unmanned Ground Vehicles (UGVs) \cite{krishna1997tethering, schempf2009self}. Most Remotely Operated Vehicles (ROVs) for underwater exploration are also tethered \cite{yuh2000design}. Both tethered UGVs and UMVs didn't address the tether specifically and assumed that the movement of the tether happens in completely free space and wouldn't cause any problems. 

One important advantage brought in by a tether to UAVs is an alternative localization and navigation methodology, which would impact implementation of a tether planner. In lieu of traditional UAV sensory inputs including GPS \cite{oh2015extended}, Inertia Measurement Units (IMUs) \cite{grzonka2012fully, achtelik2009autonomous, carrillo2012combining}, laser range finder \cite{grzonka2012fully, achtelik2009autonomous}, stereo \cite{achtelik2009autonomous, carrillo2012combining} and RGB-D cameras \cite{vetrella2015rgb}, the new approach comes with a tether-based UAV controller: tether length encoder, tether azimuth and elevation angle computed by fusing onboard IMU and tether angle sensor \cite{lupashin2013stabilization}. The localization of the UAV could then be realized by measuring the tether length and angles with respect to the global frame. Work in \cite{lupashin2013stabilization} is most relevant to this research, since they dealt with localization and stabilization of the UAV with a taut tether. This laid the ground work for this paper since our low level UAV position and attitude controller is based on a taut tether. However, the UAV in \cite{lupashin2013stabilization} was assumed to be hovering in free space without any obstacles. The tether was taut and straight, not in contact with the environment. This worked well in outdoor environments and free indoor spaces. But in more realistic scenarios it is not reasonable to always assume a free-flight zone, where the tether never touches the environment. Tether contact has been researched in the sense of tether entanglement prevention in multi-robot systems \cite{hert1999motion}. But researchers have not paid sufficient attention to the interaction of tether with physical environments, especially in 3-D cluttered spaces. 

This paper looks into how to generate executable motion plans for a tethered UAV in 3-D cluttered environments, for purposes such as visual assisting of a primary ground robot. Two approaches are proposed: (1) using a ray-casting method to reduce the reachable space of the UAV, assuming a straight and non-contact tether and (2) integrating tether contact points into the planning space so the UAV could work with a kinked tether.

\section{Approach}
\label{sec::approach}
Two motion planners and a motion executor are discussed in this section. It is assumed that a map is precomputed in the form of 3-D occupancy grid. The static and complete map is occupied with obstacles. The goal is to generate an executable path for a tethered UAV from point \textit{start} to point \textit{goal} in the free space. The path is then interpreted and translated to tether-based motion commands. Reachable space reduction via ray casting assumes the tether does not contact the obstacles during the whole flight. Contact point(s) planning and relaxation allows the tether to touch the obstacles and form a kink, and therefore the UAV can reach any free space as if it were tetherless. After the offline motion planners generate waypoints with contacts in the given 3-D map, the online motion executor issues real-time motion commands based on the motion plan and tether localization feedback. 

\subsection{Reachable Space Reduction via Ray Casting}
In contrast to conventional UAVs, tethered UAVs have to maintain connection with its ground station via a tether. If the tether needs to remain taut and straight, no contacts are allowed with the environments. This constraint reduces the reachable space. Obstacles cannot locate between the UAV and its tether reel, since otherwise the tether in between would touch the obstacles. Based on this idea, this planner uses a ray casting approach from tether reel to obstacles in order to identify spaces in the configuration space, which are feasible for the UAV alone, but not with a tether. For voxels on the ray, those between reel center and obstacles are still open, while those beyond obstacles are blocked. 

After reachable space reduction, remaining space is completely free even with respect to the tether. Any path planning algorithm which works in 3-D could be applied between any two points in the reduced space. Here, Probabilistic Road Map (PRM) \cite{kavraki1996probabilistic} is used to plan an executable path in the reduced space. 

\begin{figure}[!t]
\centering
\includegraphics[width=0.9\columnwidth]{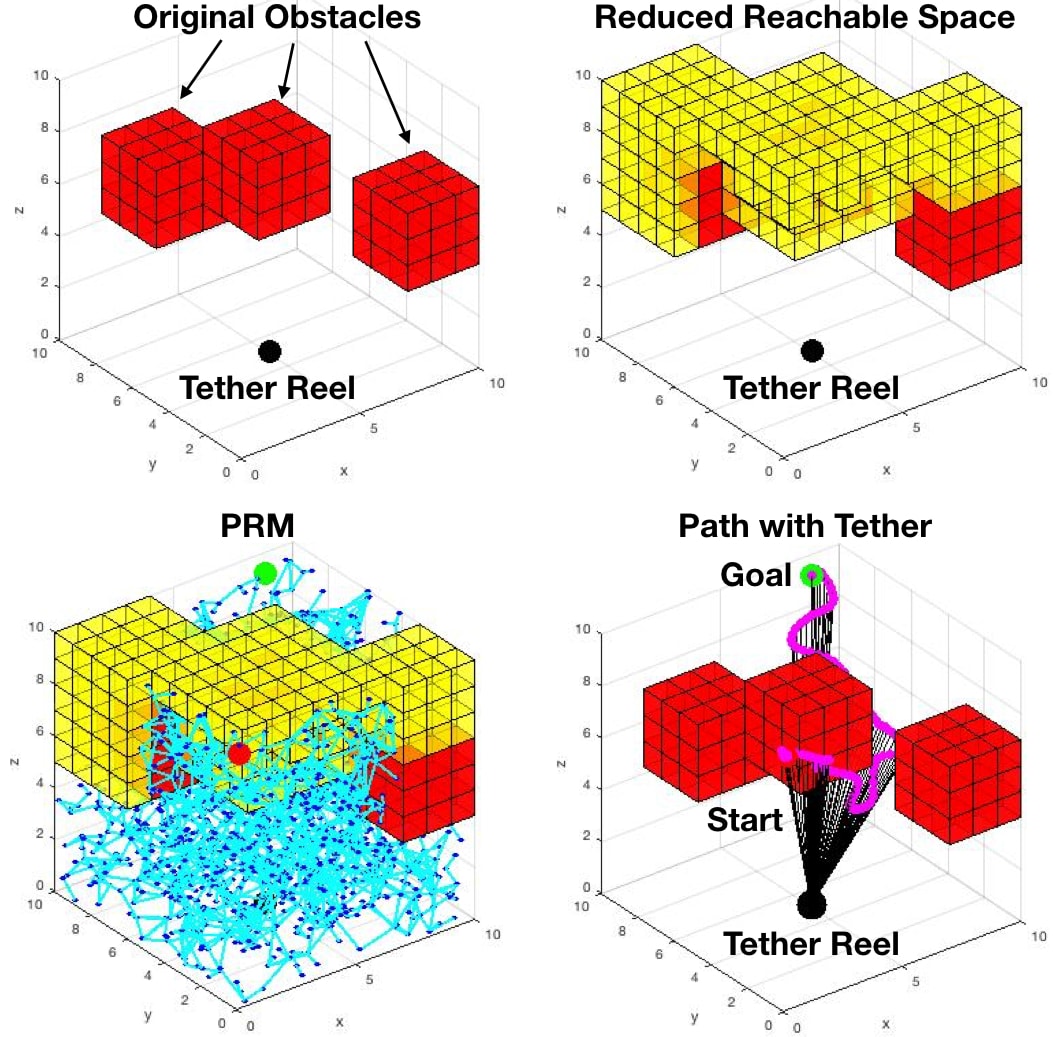}
\caption{Reachable space is reduced by ray casting from tether reel to original obstacles (in red). Yellow voxels are non-reachable space due to tether. UAV path is planned using PRM (cyan) in the reduced reachable space. The UAV has to go beneath the obstacles since a direct straight path above all obstacles is not allowed by the existence of tether. }
\label{fig:ray_casting}
\end{figure}

Ray casting has a complexity proportional to the number of obstacles $\mathcal{O}(o)$. Assuming $n$ is the number of vertices in the road map, the complexity of PRM is $\mathcal{O}(n^2)$, including populating free space, running local planner, and final search to find a path. The complete algorithm pipeline with complexity $\mathcal{O}(o+n^2)$ is illustrated in Fig. \ref{fig:ray_casting}. 

\subsection{Contact Point(s) Planning and Relaxation}
As we can see in Fig. \ref{fig:ray_casting}, the free space is largely reduced by ray casting, leaving only a subset of the original free space reachable with a tether. A different algorithm with tether contact point(s) planning is presented here. It automatically plans the tether contact point(s) when the robot locates in the originally free but actually occupied spaces after reduction (yellow voxels in Fig. \ref{fig:ray_casting}). It also has the capability to relax the contact point(s) when the UAV returns to the post-reduction free space. It assumes that once a contact point is formed, it doesn't move unless being relaxed. The algorithm is outlined in Algorithm \ref{alg::contact_planning}.

\begin{algorithm}[!t]
 \caption{Contact Point(s) Planning and Relaxation}
 \begin{algorithmic}[1]
 \renewcommand{\algorithmicrequire}{\textbf{Input:}}
 \renewcommand{\algorithmicensure}{\textbf{Output:}}
 \REQUIRE \textit{map}, \textit{start}, \textit{goal}, \textit{tether\_origin}
 \ENSURE  executable path: waypoints with contact points
  \STATE Inflate original \textit{map}
  \STATE Generate PRM in free space
  \STATE Query and smooth path from \textit{start} to \textit{goal} 
  \STATE Initialize \textit{CP\_stack} with \textit{tether\_origin}
  \STATE Attach \textit{tether\_origin} to all waypoints \textit{WP}s on path
  \STATE \textit{relax\_flag} = 0 
  \FOR {every \textit{WP} on path}
  \STATE \textit{curent\_contact} = \textit{CP\_stack} top \textit{CP}
  \IF {(\textit{CP\_stack} has more than one CPs)}
  \STATE \textit{last\_contact} = second \textit{CP} from \textit{CP\_stack} top
  \STATE \textit{collision\_flag} = CheckCollision (\textit{last\_contact}, \textit{WP}, \textit{map})
  \IF {\textit{collision\_flag} == 0}
  \IF {ObstacleConfined (\textit{curent\_contact}, \textit{last\_contact}, \textit{WP}, \textit{map})}
  \STATE \textit{relax\_flag} = 0
  \ELSE
  \STATE \textit{relax\_flag} = 1 // contact relaxation
  \ENDIF
  \ELSIF {\textit{collision\_flag} == 1} 
  \STATE \textit{relax\_flag} = 0
  \ENDIF
  \ENDIF
  \IF {\textit{relax\_flag} == 1}
  \STATE pop \textit{CP\_stack}
  \STATE attach new \textit{CP\_stack} top to all following \textit{WP}s
  \STATE  \textit{relax\_flag} = 0
  \ELSIF {\textit{relax\_flag} = 0}
  \IF {CheckCollision (\textit{current\_contact}, \textit{WP}, \textit{map})}
  \STATE push new \textit{CP} to \textit{CP\_stack} // contact planning 
  \STATE attach new \textit{CP\_stack} top to all following \textit{WP}s
  \ENDIF
  \ENDIF
  \ENDFOR
 \RETURN all \textit{WP}s along with their \textit{CP}s
 \end{algorithmic}
 \label{alg::contact_planning}
 \end{algorithm}

\begin{figure*}[!t]
\centering
\subfloat[Original Configuration Space with Current and Last Contact Points]{\includegraphics[width=0.485\columnwidth]{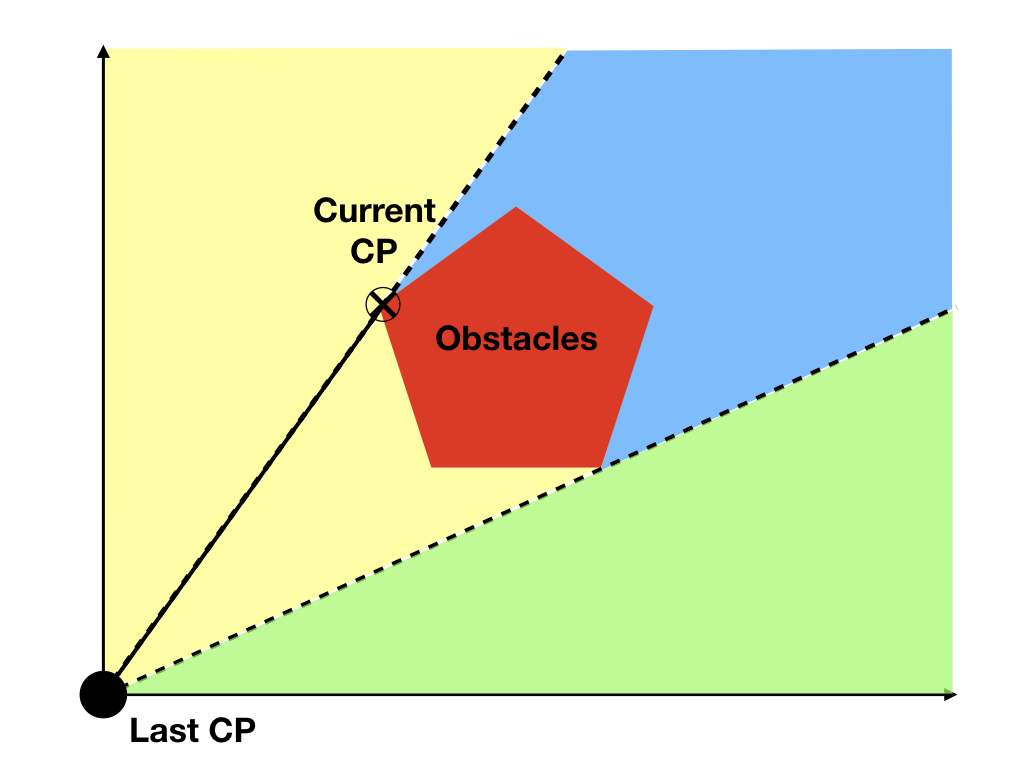}%
\label{fig::original}}
\hspace{0.015\columnwidth}
\subfloat[Current Contact Point Relaxed due to No Collision and Obstacles Not Being Confined]{\includegraphics[width=0.485\columnwidth]{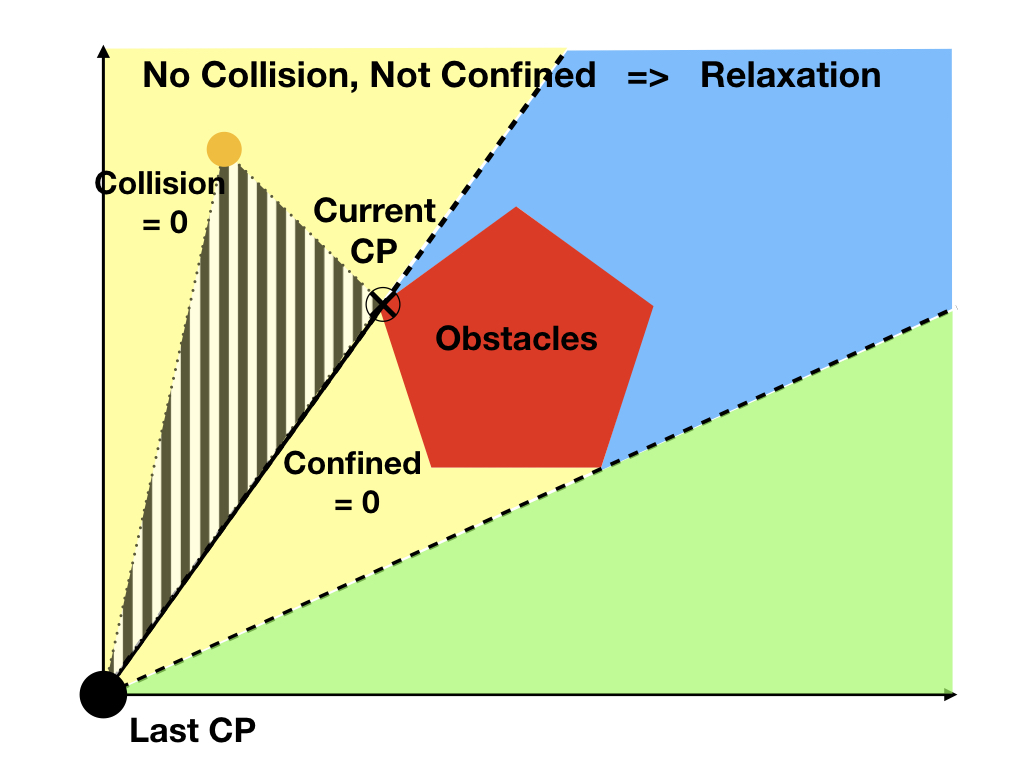}%
\label{fig::relaxation}}
\hfil
\subfloat[Current Contact Point Not Relaxed due to No Collision and Obstacles Being Confined]{\includegraphics[width=0.485\columnwidth]{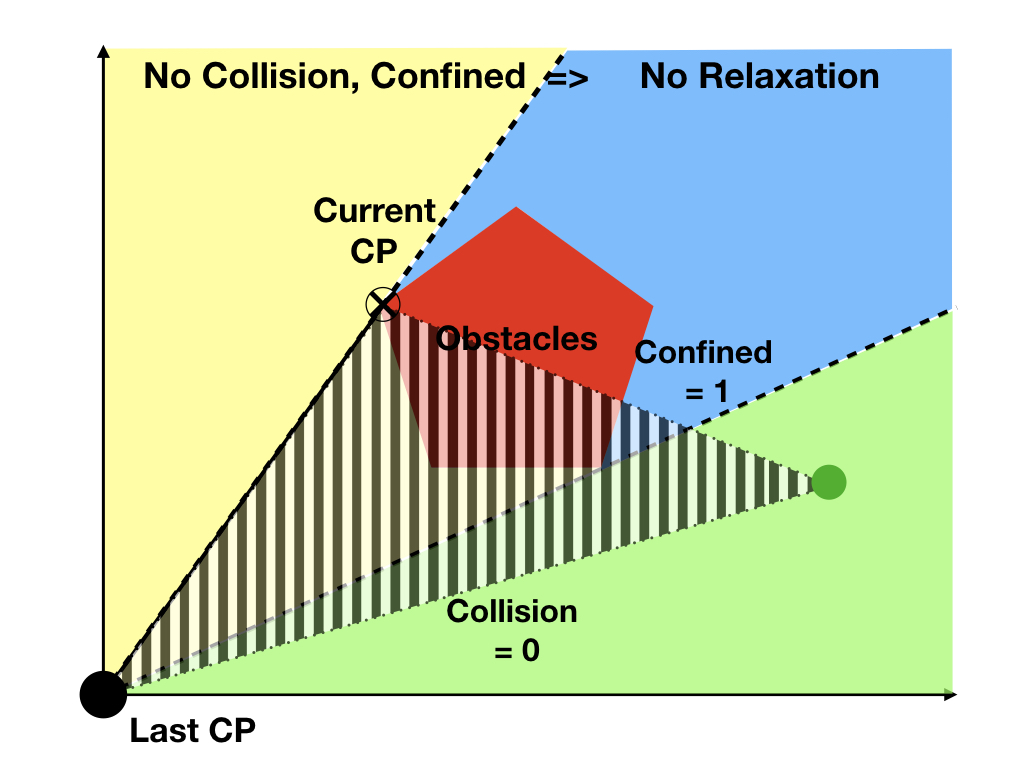}%
\label{fig::no_relaxation1}}
\hspace{0.015\columnwidth}
\subfloat[Current Contact Point Not Relaxed due to Collision]{\includegraphics[width=0.485\columnwidth]{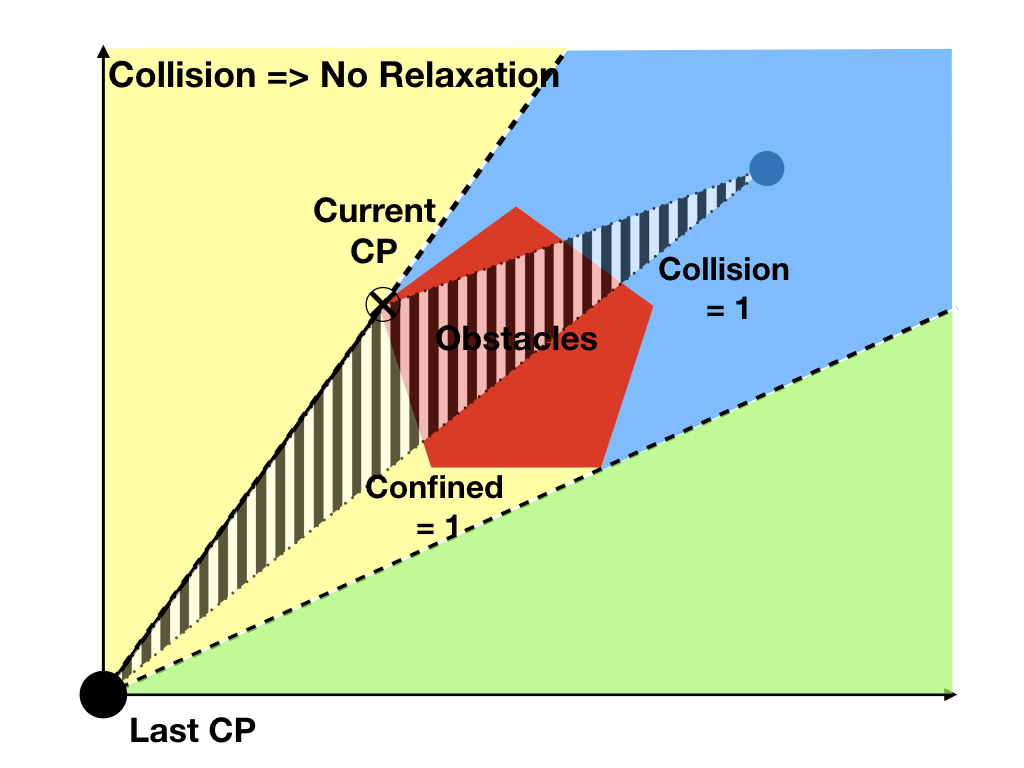}%
\label{fig::no_relaxation2}}
\caption{2-D Representation of the Tether Relaxation Scheme: based on \textit{CheckCollision} between last contact point and current waypoint with the map, \textit{ObstacleConfined} checks if any obstacles are confined within the triangle formed by waypoint, last and current contact points.}
\label{fig:relaxation}
\end{figure*}

The algorithm starts with inflating the original \textit{map} with the radius of the robot, generating PRM in free spaces, and query a  smooth path from \textit{start} to \textit{goal}. We use a stack (\textit{CP\_stack}) to keep track of all current active tether contact points with the environment. Line 5 initializes contact points of all waypoints along the path to be the original tether reel center (\textit{tether\_origin}). \textit{relax\_flag} indicates if contact point relaxation is necessary. The rest of the algorithm plans the contact point for each individual waypoint. Line 9 to 21 determines if it is necessary to relax the current contact point. It first checks whether the robot is located at a waypoint directly reachable from the last contact point (line 11). If true (\textit{collision\_flag} == 0), it is possibly necessary to relax the current contact point (\textit{CP}), depending on if obstacle is confined within the triangle formed by the waypoint, last and current contact points. If false (\textit{collision\_flag} == 1), relaxation is not necessary. The actual relaxation is implemented in line 22 to 25. The current \textit{CP} is popped from \textit{CP\_stack}, and the last contact point is assigned to all subsequent waypoints. If relaxation is not necessary, line 27 checks if it's necessary to form a new contact point. If yes, the new \textit{CP} is pushed into \textit{CP\_stack} and all following waypoints are assigned the new contact point. 

\textit{CheckCollision (point A, point B, map)} draws a line between \textit{point A} and \textit{Point B} and see if any points on the line intersect with any obstacles in \textit{map}. If there is no collision, \textit{ObstacleConfined (point A, point B, point C, map)} further checks if any obstacle in \textit{map} is confined in the triangle formed by \textit{point A}, \textit{point B}, and \textit{point C}. A 2-D illustration of the tether relaxation pipeline is shown in Fig. \ref{fig:relaxation}. The 3-D version works on the projection onto x-y, y-z, and x-z planes. To be 3-dimensionally confined, obstacle needs to be 2-dimensionally confined in all three projection planes. 

Both \textit{CheckCollision} and \textit{ObstacleConfined} have a complexity proportional to the number of obstacles $\mathcal{O}(o)$. The complexity for PRM is still $\mathcal{O}(n^2)$. Assuming the executable path consists of $p$ waypoints, the whole algorithm's complexity is $\mathcal{O}(n^2+po)$. 

The result of the algorithm is an executable path composed of 3-D waypoints along with corresponding 3-D contact points. If the tether is not touching the environment, contact point is treated as the tether reel center. So the motion planner outputs a 6-D waypoints and contact points file.  

\subsection{Motion Executor}
The offline 6-dimensional motion plan is parsed by the online motion executor. The UAV is commanded to reach every single waypoint along the path. In this work, we treated the UAV as a mass point and thus only consider positional movement. The vehicle position control uses \emph{tether length} $r$, \emph{elevation} $\theta$ (vertical angle of tether), and \emph{azimuth} $\phi$ (horizontal angle of tether). The position of the vehicle could be represented in polar coordinate system (Fig. \ref{fig:executor}). Given a certain $x$, $y$, and $z$, $r$, $\theta$, and $\phi$ could be easily derived: 

\begin{figure}[!t]
\centering
\includegraphics[width=0.9\columnwidth]{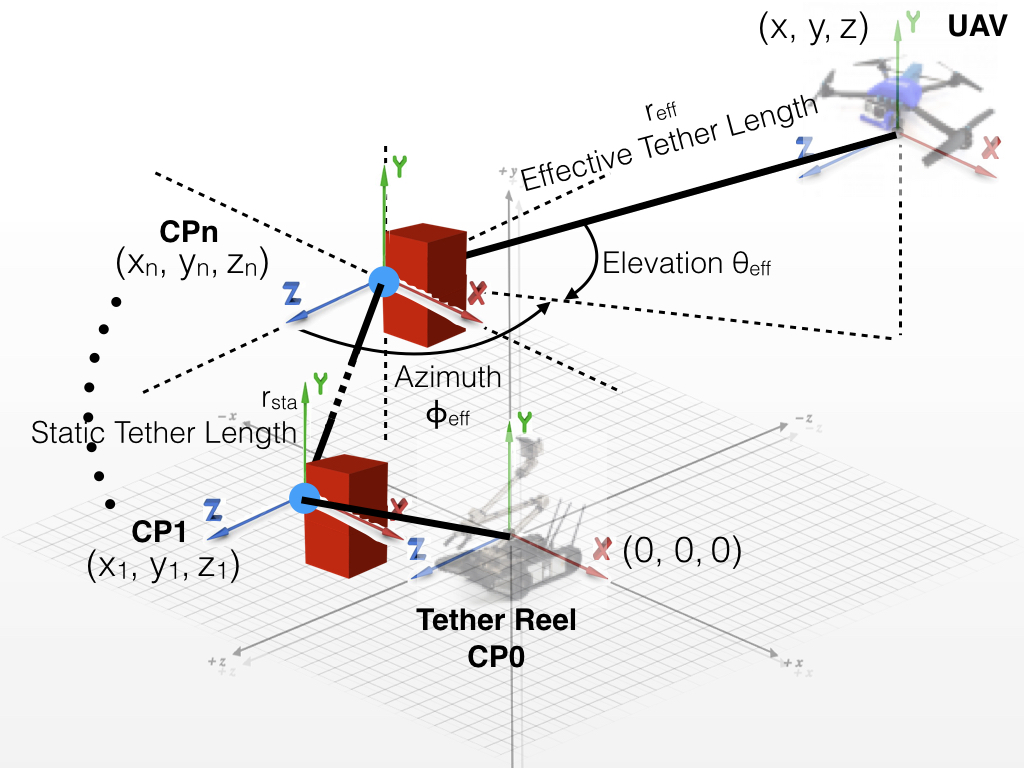}
\caption{Motion Executor Interpretation: Tether contact points are saved in a stack, where the latest contact point locates at the top. Tether is divided into several straight line segments, whose lengths are saved and associated with each contact point. Positional control is based on the planned relative coordinates of the UAV with respect to the last contact point. }
\label{fig:executor}
\end{figure}


\begin{equation}
\label{eqn::euclidean2polar}
\left\{\begin{matrix}
r = &\sqrt{x^2+y^2+z^2}\\
\theta = &arcsin(\frac{y}{\sqrt{x^2+y^2+z^2}})\\
\phi = &atan2(\frac{x}{z})
\end{matrix}\right.
\end{equation}

However, since in this work multiple contact points are allowed,  all three control parameters, $r$, $\theta$, and $\phi$, are not relative to the tether reel (origin), but to the last contact point $CP_n$. Here, we use a stack to store all contact points. Whenever the motion executor reaches a new contact point, it pushes it into the stack. It also saves the current static tether length ($r_{sta}$) from the reel to this contact point. It is termed as static since this portion of the tether remains static based on our assumption that formed contact points don't move unless being relaxed. Whenever a contact point is relaxed, the motion executor pops it from the stack and reduces the static tether length by the corresponding segment length. So we have:

\begin{equation}
\label{eqn::static_tether_length}
r_{sta} =  \sum_{0}^{n-1}\sqrt{(x_{i+1}-x_i)^2+(y_{i+1}-y_i)^2+(z_{i+1}-z_i)^2}
\end{equation}

Since we have all our controls with respect to the last formed contact point (top of stack), we have effective values relative to this point: 

\begin{equation}
\label{eqn::effective_controls}
\left\{\begin{matrix}
r_{eff} = &\sqrt{(x-x_n)^2+(y-y_n)^2+(z-z_n)^2}\\
\theta_{eff} = &arcsin(\frac{y-y_n}{\sqrt{(x-x_n)^2+(y-y_n)^2+(z-z_n)^2}})\\
\phi_{eff} = &atan2(\frac{x-x_n}{z-z_n})
\end{matrix}\right.
\end{equation}

So the desired controls are: 
\begin{equation}
\label{eqn::desired_controls}
\left\{\begin{matrix}
r = &r_{eff} + r_{sta}\\
\theta = &\theta_{eff}\\
\phi = &\phi_{eff}
\end{matrix}\right.
\end{equation}

The desired values of $r$, $\theta$, and $\phi$ are regulated by a PID controller based on the sensory feedback from the UAV (tether angle sensors and reel encoder). An acceptance radius $R_{acc}$ is defined so that whenever the UAV reaches a ball with radius $R_{acc}$ around the desired waypoint, this waypoint is treated as reached and the executor moves on to the next waypoint. 

The motion executor doesn't need to discriminate between two different motion planners. The waypoint file from the ray casting approach could also be 6-dimensional, with all contact points to be the tether reel, namely the origin of the global coordinate system. This also applies to the non-contact path segment(s) from the contact planning approach. 

\section{Exploratory Trials}
\label{sec::experiments}

The purpose of the exploratory trials is proof of concept of our two motion planning algorithms: reachable space reduction by ray casting and contact point(s) planning and relaxation. By running experiments on physical robots, we wanted to show that our motion planners can navigate the UAV between two points in the corresponding free space of each planner. The trial completion was determined based on the UAV's onboard localization. By running our two motion planning algorithms, we also wanted to demonstrate different reachability sets achievable by the two planners. It was computed as a percentage of reachable spaces in the whole map by offline computation based on the obstacles in the map. Finally, navigation accuracy in terms of cross track error was presented by comparison between planned paths and executed paths. The latter was captured by a ground truth motion capture (MoCap) system. 

\begin{figure}[!t]
\centering
\includegraphics[width=1\columnwidth]{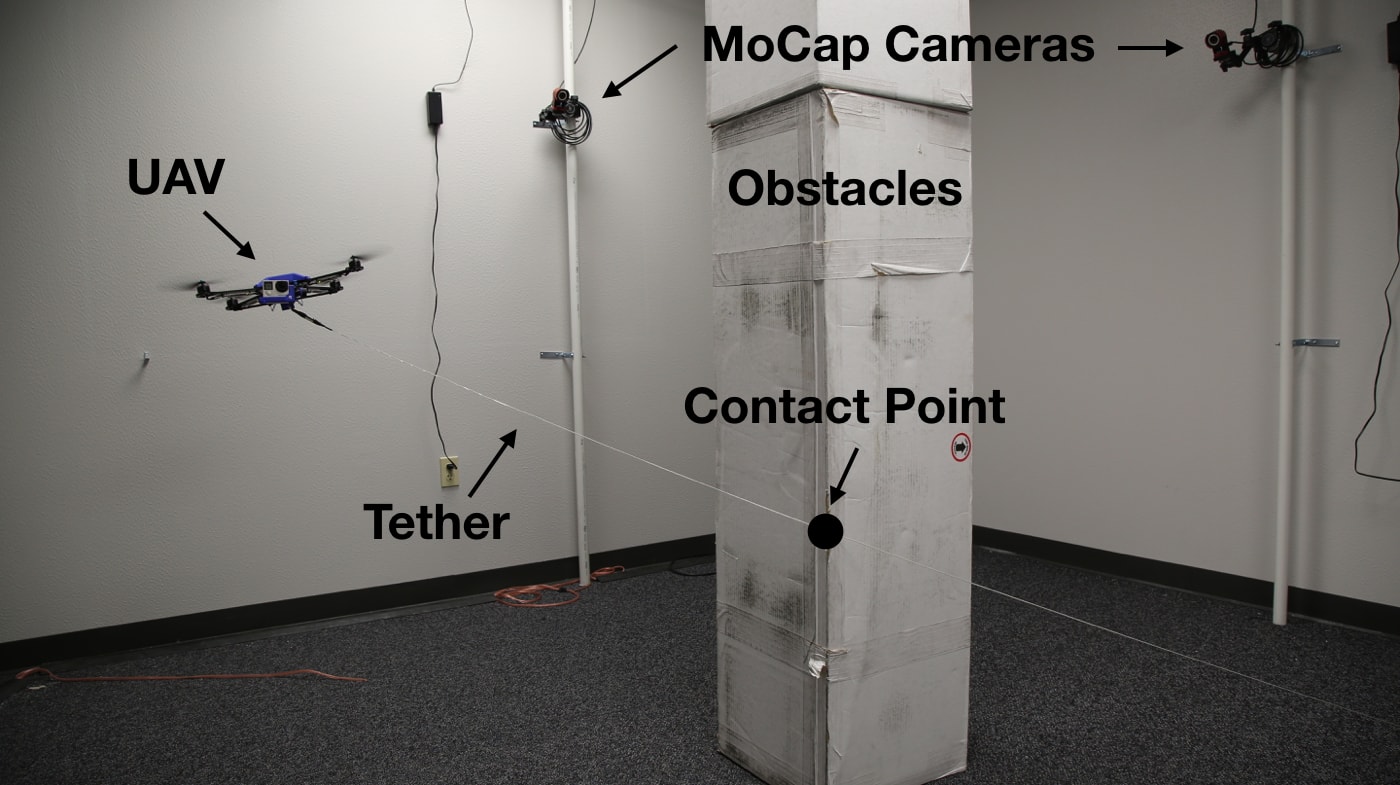}
\caption{UAV flying with one tether contact point in the MoCap studio}
\label{fig:mocap}
\end{figure}

Our exploratory trials were conducted in a motion capture studio to capture motion ground truth. The studio is equipped with 12 OptiTrack Flex 13 cameras running at 120 Hz. The 1280$\times$1024 high resolution cameras with a 56\degree~Field of View provide less than 0.3mm positional error and cover the whole 3.3$\times$3.3$\times$2.97m space. The high number of cameras guarantee that the UAV could be captured even if the markers were blocked by the obstacles from some cameras. We used obstacles made of cardboard, which formed a 0.33$\times$0.33$\times$0.297m vertical shaft and located in the middle of the experimental environment. The choice of cardboard was to guarantee safe tether contact. This configuration of obstacles blocked most direct passages between different regions in the map, and was particularly difficult for a tethered UAV to navigate through. Fotokite Pro was used as our tethered UAV. The online motion executor executed the offline motion plan from the two algorithms. During the physical tests, the acceptance radius $R_{acc}$ was set to 0.4m. This is the best localization accuracy achievable by Fotokite's sensory feedback measured by experiments. Fig. \ref{fig:mocap} shows the tethered UAV flying in the MoCap studio. 

\begin{figure*}[!t]
\centering
\subfloat[Reduced Reachable Space by Ray casting]{\includegraphics[width=0.66\columnwidth]{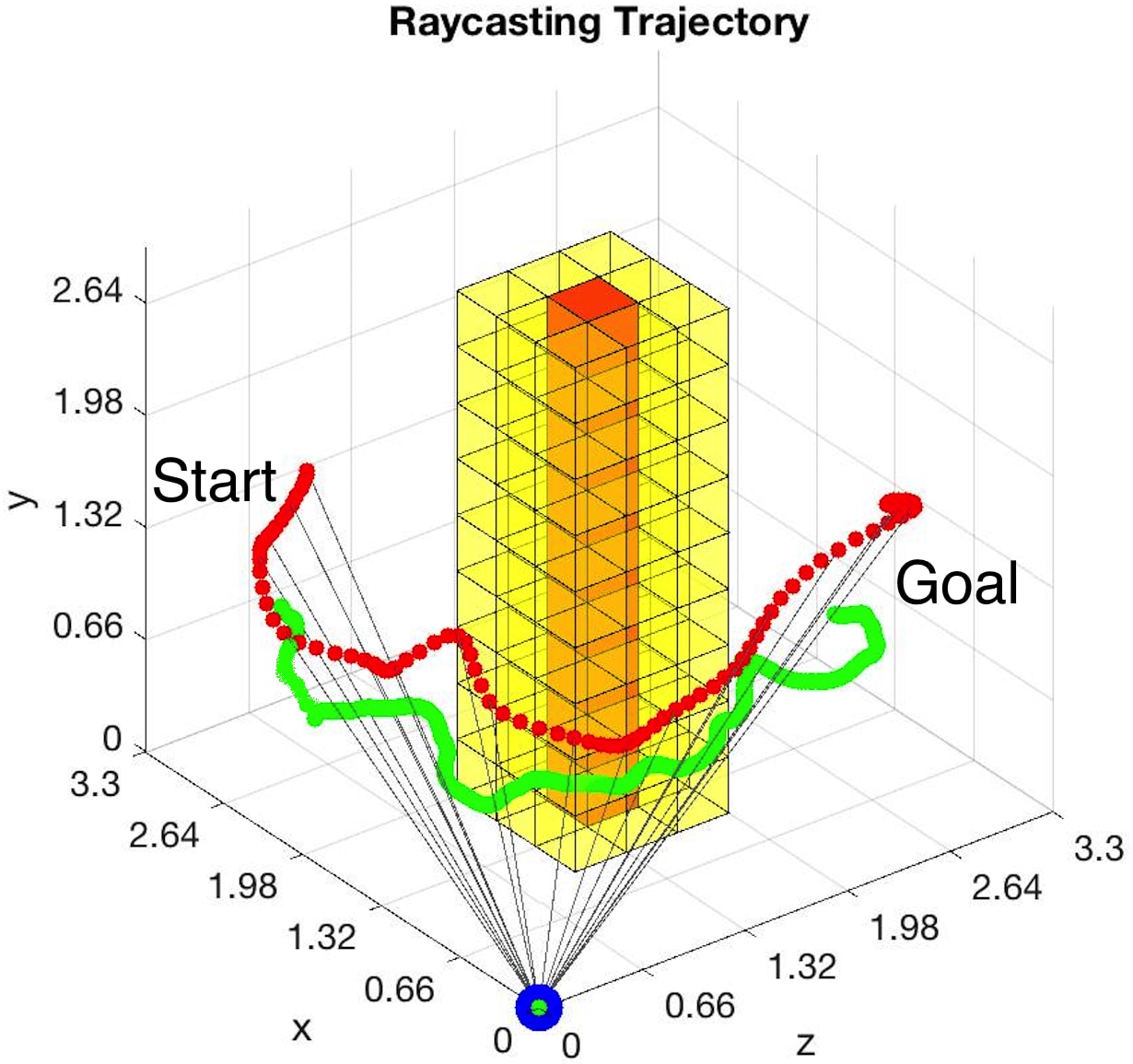}%
\label{fig::raycasting_traj}}
\subfloat[Contact Point with Relaxation]{\includegraphics[width=0.66\columnwidth]{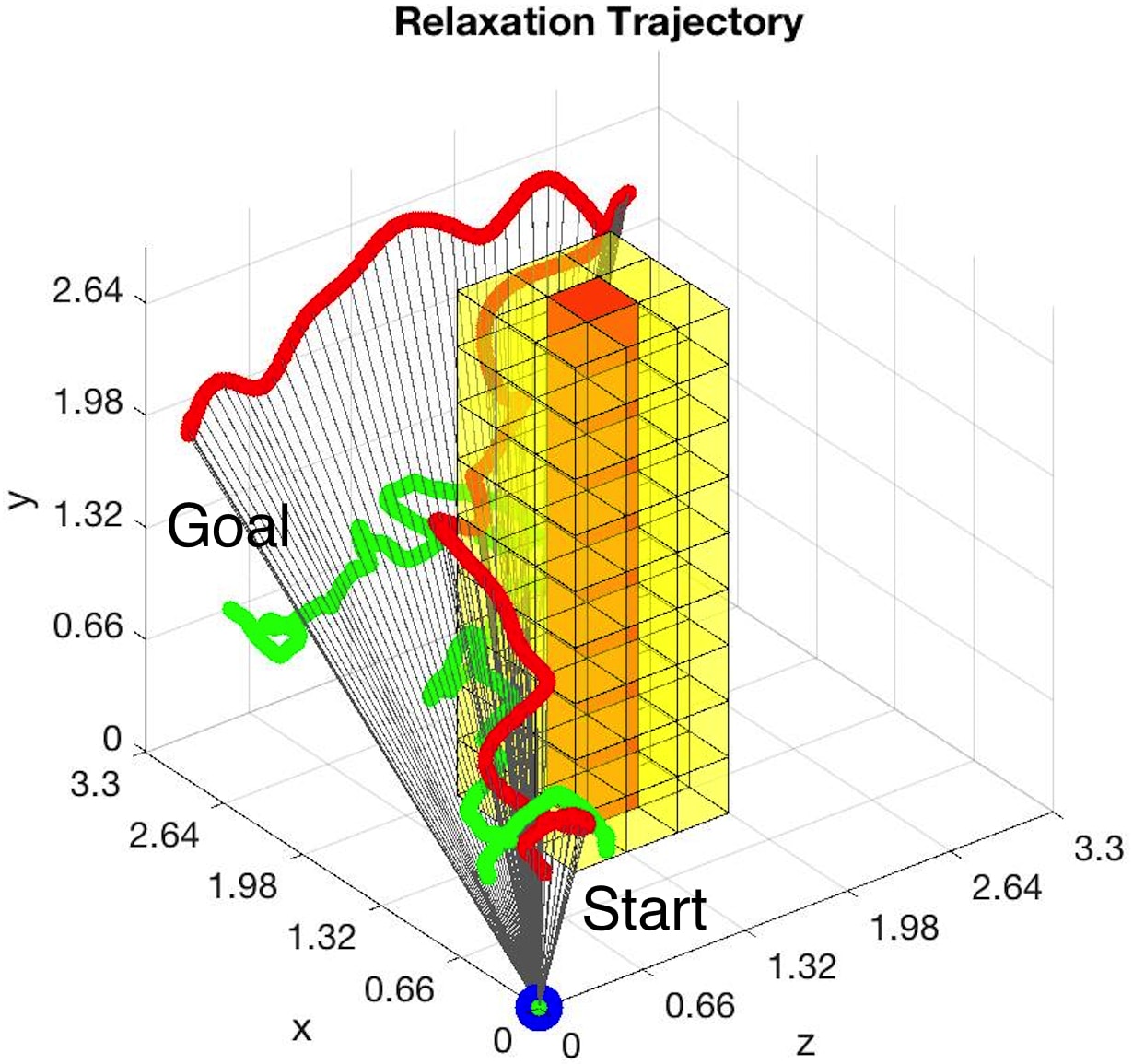}%
\label{fig::relaxation_traj}}
\subfloat[Two Contact Points]{\includegraphics[width=0.66\columnwidth]{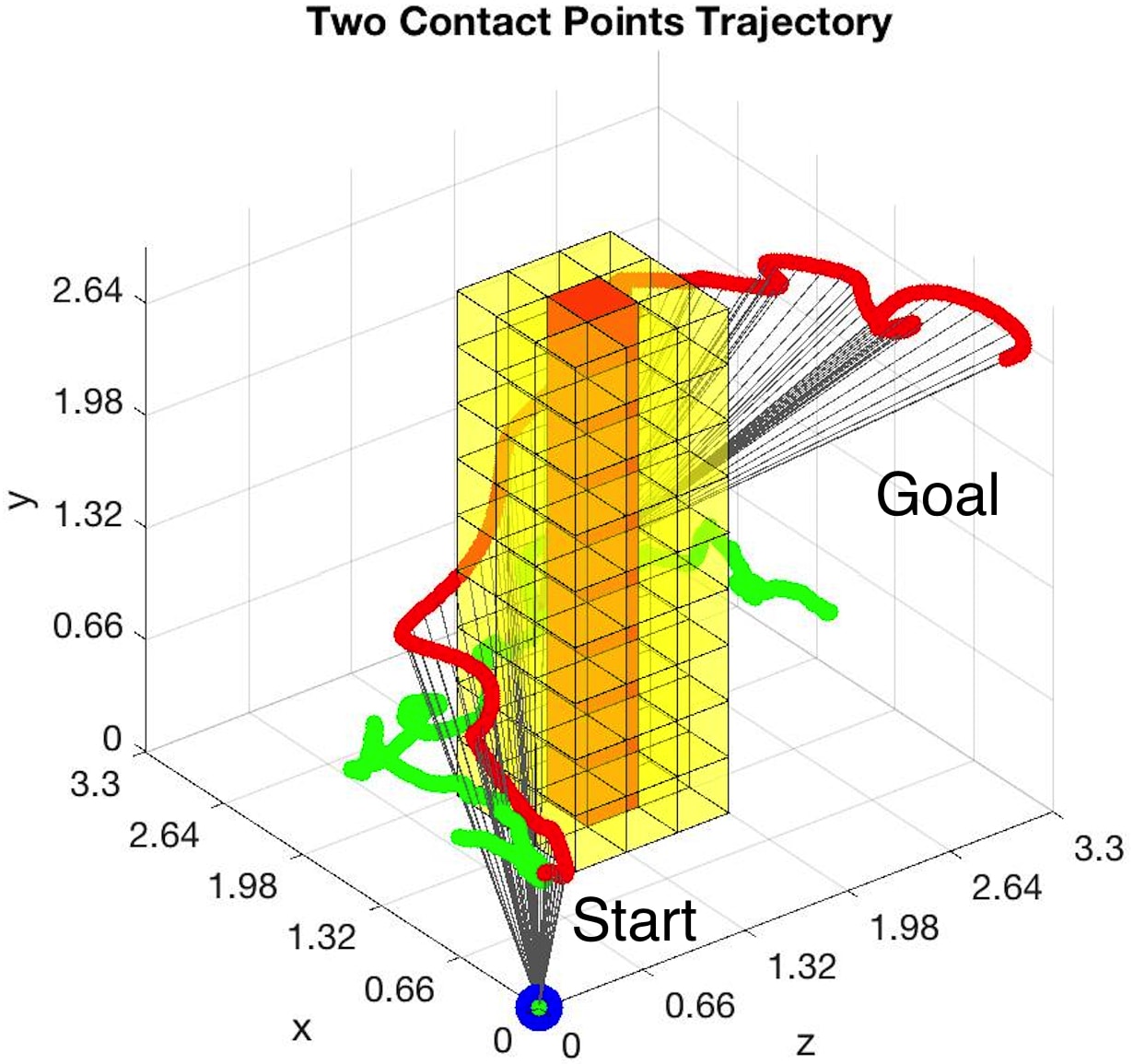}%
\label{fig::two_contact_traj}}
\caption{Three Different Paths Planned (Red) and Executed (Green): Red voxels represent the obstacles and yellow voxels are the occupied spaces due to map inflation. Red path is the off-line computed motion plan and green one is the actual path captured by OptiTrack motion capture system.}
\label{fig::results}
\end{figure*}

In order to validate our motion planning algorithms, we conducted three sets of experiments on the tethered UAV: 

\begin{itemize}
\item Moving in free space after reachable space reduction using ray casting (Fig. \ref{fig::raycasting_traj})
\item Returning to free space by relaxing previously formed contact point (Fig. \ref{fig::relaxation_traj})
\item Entering non-reachable space with a straight tether by planning two contact points (Fig. \ref{fig::two_contact_traj})
\end{itemize}

Since the two different motion planners are dealing with different configuration spaces, i.e. reduced and original reachable spaces, we cannot replicate the same navigation task (same \textit{start} and \textit{goal}) for both of them. For the first set of experiments, we manually chose pairs of \textit{start} and \textit{goal} in the reduced reachable space. The obvious direct paths between the pairs were not executable due to the tether. The ray casting motion planner needed to come up with an alternative path to circumvent the obstacles to remain a straight tether. For the second set of experiments, we manually chose a tuple of \textit{(start, middle point, goal)} in the original free space. The \textit{middle point} located at a position where one contact point was necessary to reach. The \textit{Goal} located at a position where no contact was necessary. So the robot had to form and then relax the contact point to reach the final target during the flight. For the third set of experiments, we manually chose a tuple of \textit{(start, middle point, goal)}. The \textit{middle point} located at a position where one contact point was necessary to reach. The \textit{Goal} located at a position where two contacts were necessary. So the robot had to form two contact points in a row to reach the final target during the flight. 

Based on the given map, we obtained two different reachable spaces from the two motion planners. The ray casting method reduced navigable space from the original free space while contact(s) point planning and relaxation kept the whole free space intact. 


We totally performed 21 trials. Two trials were discarded due to the UAV platform hardware failure and one was discarded due to the UAV flying out of the range of the MoCap system. We obtained 18 planned paths with way points and contact points (CPs for ray casting were simply tether reel) and corresponding 18 executed paths captured at 120Hz, six trials for each set.

\section{DISCUSSION}
\label{sec::discussions}
\subsection{Comparison of the Two Algorithms}
Ray casting works in post-reduction free spaces and the UAV cannot reach spaces blocked by ray casting. Contact point(s) planning can navigate to spaces which are not reachable with a straight tether. Tether can be properly relaxed when UAV returns to original free spaces. Multiple contact points could be formed and handled. For this particular set up, ray casting can reach 60\% of the whole free space, and contact point(s) planning can reach 100\%. A 40\% reduction of reachable space was observed for ray casting to maintain a straight tether. Contact point planning has greater reachability since the UAV is de facto tetherless, but tether contact may not be acceptable in all domains. There is an open issue as to whether the tether would break or would damage the environment. Ray casting has a complexity of $\mathcal{O}(o+n^2)$. Contact point(s) planning has $\mathcal{O}(n^2+po)$ due to the extra work load to plan and relax contact points. 


\subsection{Insights on Implementation}
All 18 trials were completed based on the UAV sensor feedback. However, the onboard localization error accumulates during flight, so position estimation is not precise. Fig. \ref{fig::results} shows three example trials. Fig. \ref{fig::raycasting_traj} shows the execution of the path generated by ray casting method. Fig. \ref{fig::relaxation_traj} and Fig. \ref{fig::two_contact_traj} use contact point(s) planning and relaxation. To be noticed is that the tether can pass through yellow voxels (inflation) and contact points can only be formed on the surfaces/edges of the red voxels (obstacles). In Fig. \ref{fig::raycasting_traj}, originally free spaces behind the obstacles are blocked by ray casting, so the robot has to forgo the short path behind the obstacles and circumvent from the front in order to maintain a straight tether through the whole flight. Fig. \ref{fig::relaxation_traj} shows the robot firstly navigates to the far end of the map, where the tether has to touch the obstacles. One contact point is planned, which is thereafter relaxed, since the robot flies back to the non-contact space and reaches the final destination with a straight tether. As we can see, the navigation accuracy decreases significantly after making the contact. In Fig \ref{fig::two_contact_traj}, two contact points are planned along the way. Although the last portion of the path is reachable directly from the tether reel, it still keeps the two contact points since obstacles are confined within the triangle formed by the waypoint, current and last contact points (Fig. \ref{fig::no_relaxation1}). 

\begin{figure}[!t]
\centering
\includegraphics[width=0.68\columnwidth]{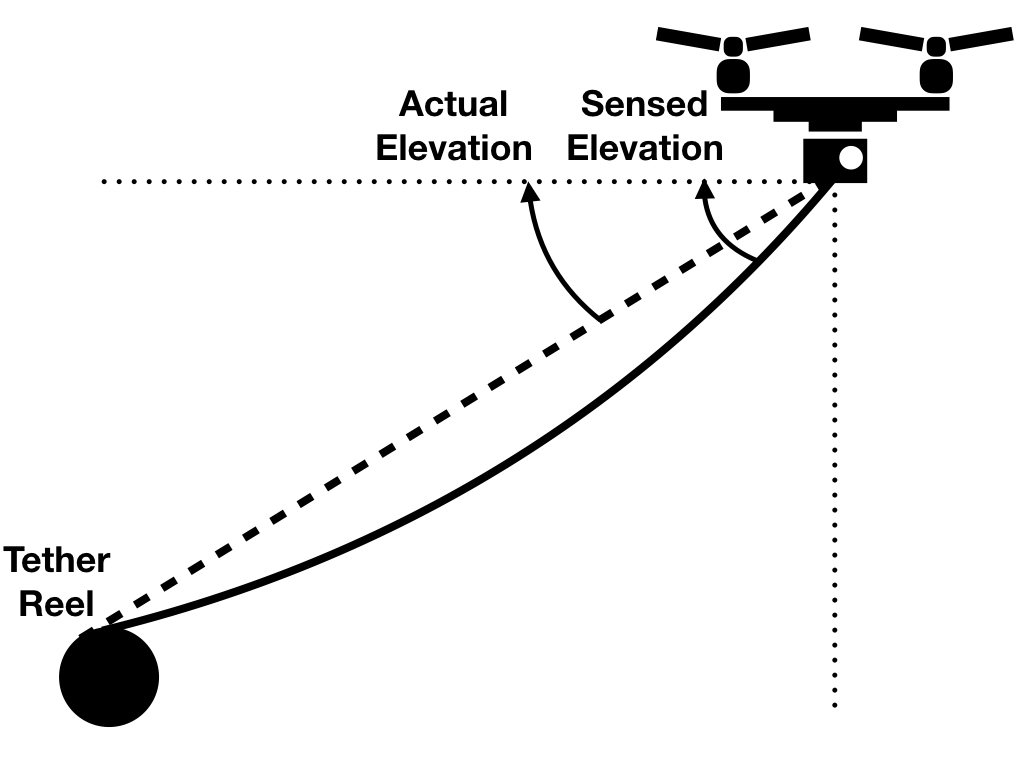}
\caption{Tethered UAV's Internal Localization Error}
\label{fig:elevation_error}
\end{figure}

One important reason behind the navigation error is UAV internal localization error, which is determined by the choice of tethered UAV. The error is further deteriorated when the UAV is far away from the tether reel. Since the increased weight of the tether will pull the tether down due to gravity, the tether will form an arc instead of a straight line (Fig. \ref{fig:elevation_error}). This explains why the height of the actual flight tends to be lower than the motion plan, since height is most relevant to the elevation angle. \cite{xiao2018indoor} investigates this effect in detail.  

\begin{figure}[!t]
\centering
\includegraphics[width=0.745\columnwidth]{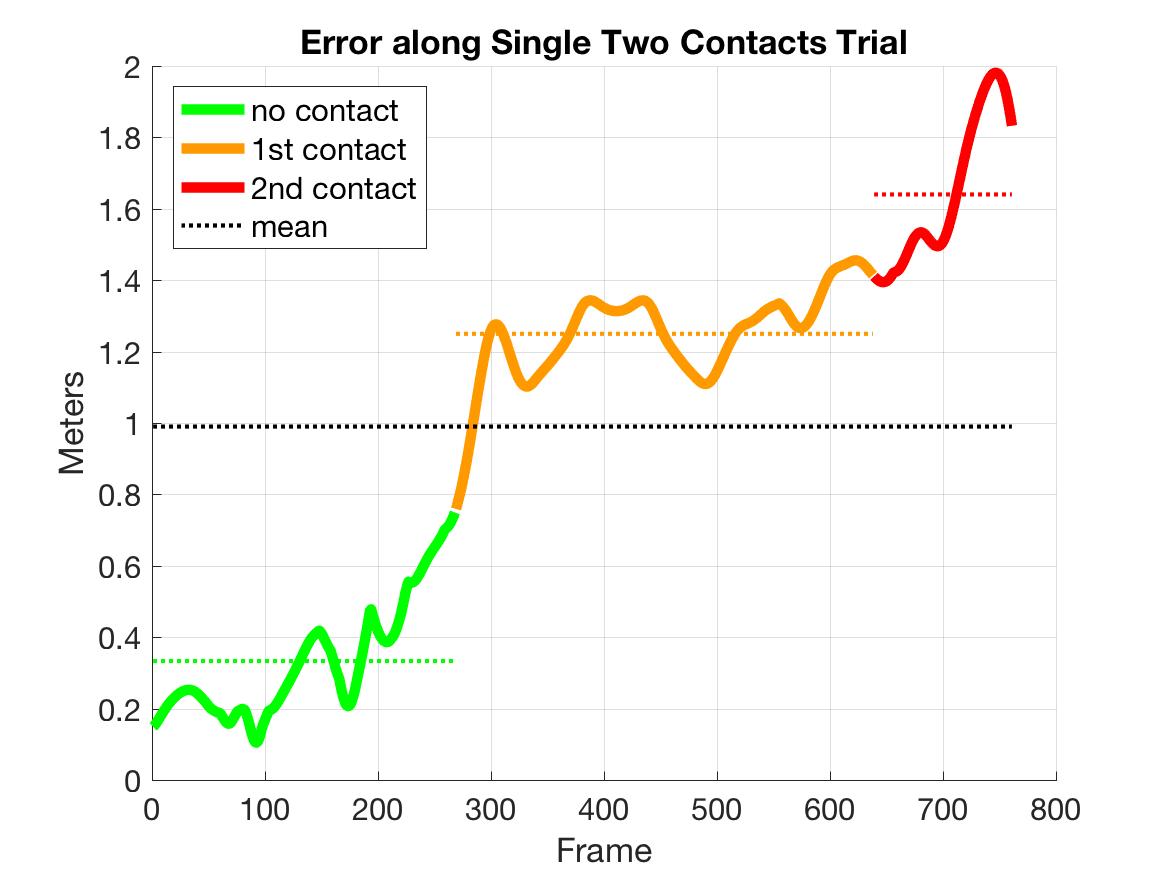}
\caption{Navigational Error along an Example Trial with Two Contact Points}
\label{fig::contact_errors}
\end{figure}

An examination on all 18 trials (Tab. \ref{tab::errors}) of the accuracy in terms of cross track error shows that contact point(s) planning has a larger error, which is even more significant with two contact points. We presume that this is due to the error introduced by contact point position, which will accumulate with increased number of contacts made. We further investigate this presumption by looking into the segmented accuracy for different contact points (Fig. \ref{fig::contact_errors}). When no contact is made, the accuracy (0.4198m) is comparable to the ray casting result in Tab. \ref{tab::errors}. The average error increases to 1.3602m at one contact and 1.7634 at two. The increased positional error is because of two reasons: (1) Due to the lack of contact point positional feedback, the actual contact point may differ from the original motion plan at initial touch. This will shift the navigation space in the next region. (2) The assumption of fixed contact point may not hold all the time, so the contact point will move slightly during flight. This process adds random noise into the system. These two sources of error will accumulate and further deteriorate the navigation accuracy. The three stages of error profile in Fig. \ref{fig::contact_errors} clearly indicate the impact of increased number of contact points: the navigational precision is less satisfactory when more contact points are formed. 

\begin{table}[]
\centering
\caption{Mean Cross Track Errors (meter)}
\label{tab::errors}
\begin{tabular}{|c|c|c|c|}
\hline
              & \textbf{Raycasting} & \textbf{\begin{tabular}[c]{@{}c@{}}1 contact \\ w/ relaxation\end{tabular}} & \textbf{\begin{tabular}[c]{@{}c@{}}2 contacts \\ w/o relaxation\end{tabular}} \\ \hline
\textbf{1}    & 0.6963              & 1.0005                                                                      & 0.9900                                                                       \\ \hline
\textbf{2}    & 0.5644              & 0.9587                                                                      & 0.9933                                                                       \\ \hline
\textbf{3}    & 0.5355              & 0.8407                                                                      & 1.1895                                                                       \\ \hline
\textbf{4}    & 0.4105              & 0.9940                                                                      & 1.1173                                                                       \\ \hline
\textbf{5}    & 0.6026              & 1.0146                                                                      & 1.1539                                                                       \\ \hline
\textbf{6}    & 0.5298              & 0.9653                                                                      & 1.1212                                                                       \\ \hline
\textbf{Mean} & 0.5565              & 0.9623                                                                      & 1.0942                                                                       \\ \hline
\end{tabular}
\end{table}


\section{CONCLUSIONS}
\label{sec::conclusions}
This paper presents and compares two motion planning methods to navigate a tethered UAV in confined spaces with obstacles. The reachable space reduction approach by ray casting provides the best navigational accuracy, but with the price of a smaller reachable space. Contact point planning allows the robot to navigate in all original free spaces as if it were tetherless. It also enables contact relaxation when necessary. However, this approach compromises motion accuracy with increased number of contact points. Two reasons were presented and errors were analyzed. The results indicate that the motion planners and executor provide an alternative way for UAV localization and navigation in indoor cluttered environments using a taut tether. They also alleviate the challenges caused by managing a tether in obstacle-occupied spaces. However, a trade-off between reachable volume and navigational accuracy exists, so full coverage of the free configuration space and high motion precision cannot be achieved at the same time. 




\section*{ACKNOWLEDGMENT}
This work is supported by NSF 1637955, NRI: A Collaborative Visual Assistant for Robot Operations in Unstructured or Confined Environments. 

\bibliographystyle{IEEEtran}
\bibliography{IEEEabrv,references}

\end{document}